\documentclass[11pt, logo, onecolumn, copyright, colorlinks=true, allcolors=blue]{nvidiatechreport}
\usepackage{graphicx} 
\usepackage{subcaption,ragged2e}
\usepackage[round]{natbib}

\usepackage[normalem]{ulem}
\usepackage[utf8]{inputenc} 
\usepackage[T1]{fontenc}    
\usepackage{url}
\usepackage{tikz}
\usepackage{enumitem}
\usetikzlibrary{positioning}
\usepackage{array}
\usepackage{caption} 
\usepackage{listings}
\usepackage{adjustbox}
\usepackage{footnote}
\usepackage{multirow}
\usepackage{amsmath}
\usepackage{amsfonts}
\usepackage{breakcites}
\usepackage{blindtext}
\usepackage{svg}
\usepackage[most]{tcolorbox}
\usepackage{tabularx}
\usepackage{array, booktabs, makecell}
\usepackage{dcolumn}
\PassOptionsToPackage{table}{xcolor}
\usepackage{siunitx}
\usepackage{caption}  
\usepackage{upquote} 

\newcommand{\nano}{\textsf{LN-Nano}\xspace}
\newcommand{\super}{\textsf{LN-Super}\xspace}
\newcommand{\ultra}{\textsf{LN-Ultra}\xspace}

\title{Llama-Nemotron: Efficient Reasoning Models}
\author{\large NVIDIA}
\date{}

\begin{document}

\begin{abstract}
\vspace{-3mm}
\large \textbf{Abstract}
\normalsize

We introduce the Llama-Nemotron series of models, an open family of heterogeneous reasoning models that deliver exceptional reasoning capabilities, inference efficiency, and an open license for enterprise use. The family comes in three sizes—Nano (8B), Super (49B), and Ultra (253B)— and performs competitively with state of the art reasoning models such as DeepSeek-R1 while offering superior inference throughput and memory efficiency. In this report, we discuss the training procedure for these models, which entails using neural architecture search from Llama 3 models for accelerated inference, knowledge distillation, and continued pretraining, followed by a reasoning-focused post-training stage consisting of two main parts: supervised fine-tuning and large scale reinforcement learning. Llama-Nemotron models are the first open-source models to support a dynamic reasoning toggle, allowing users to switch between standard chat and reasoning modes during inference.
To further support open research and facilitate model development: 
\vspace{-3mm}
\begin{itemize}
    \item We release the Llama-Nemotron reasoning models—\nano{}, \super{}, and \ultra{}—under the commercially permissive
    \href{https://www.nvidia.com/en-us/agreements/enterprise-software/nvidia-open-model-license/}{NVIDIA Open Model License Agreement}.
    \begin{itemize}
        \item \href{https://huggingface.co/nvidia/Llama-3.1-Nemotron-Nano-8B-v1}{\texttt{Llama-3.1-Nemotron-Nano-8B-v1} \includegraphics[height=0.9em]{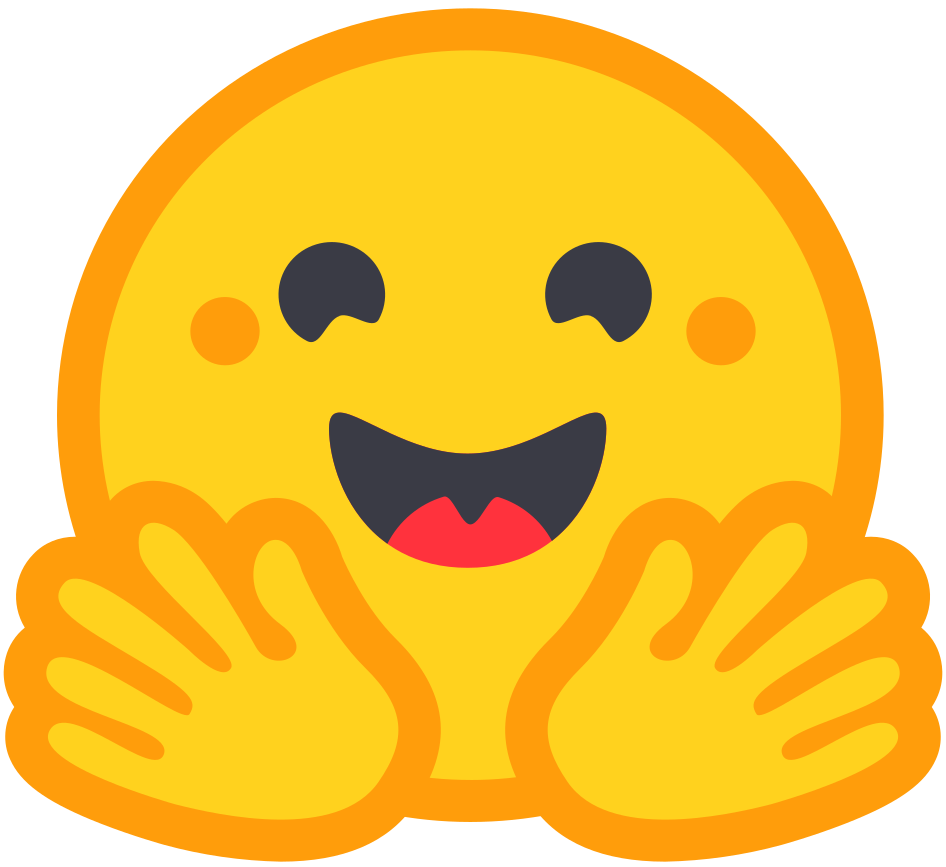}}
        \item \href{https://huggingface.co/nvidia/Llama-3_3-Nemotron-Super-49B-v1}{\texttt{Llama-3.3-Nemotron-Super-49B-v1} \includegraphics[height=0.9em]{assets/huggingface-color.png}}
        \item \href{https://huggingface.co/nvidia/Llama-3_1-Nemotron-Ultra-253B-v1}{\texttt{Llama-3.1-Nemotron-Ultra-253B-v1} \includegraphics[height=0.9em]{assets/huggingface-color.png}}
        \item \href{https://huggingface.co/nvidia/Llama-3_1-Nemotron-Ultra-253B-CPT-v1}{\texttt{Llama-3.1-Nemotron-Ultra-253B-CPT-v1} \includegraphics[height=0.9em]{assets/huggingface-color.png}}
    \end{itemize}

    \item We release the \textbf{complete post-training dataset}.
    \begin{itemize}
        \item \href{https://huggingface.co/datasets/nvidia/Llama-Nemotron-Post-Training-Dataset}{\texttt{Llama-Nemotron-Post-Training-Dataset} \includegraphics[height=0.9em]{assets/huggingface-color.png}}
    \end{itemize}

    \item We also release our training codebases: \href{https://github.com/NVIDIA/NeMo}{\texttt{NeMo}}, \href{https://github.com/NVIDIA/NeMo-Aligner/tree/llama-nemotron-dev}{\texttt{NeMo-Aligner}},
\href{https://github.com/NVIDIA/Megatron-LM}{\texttt{Megatron-LM}}.
\end{itemize}

\end{abstract}

\maketitle

\vspace{-2mm}
\begin{figure}[h]
   \centering
   \includegraphics[width=0.95\linewidth]{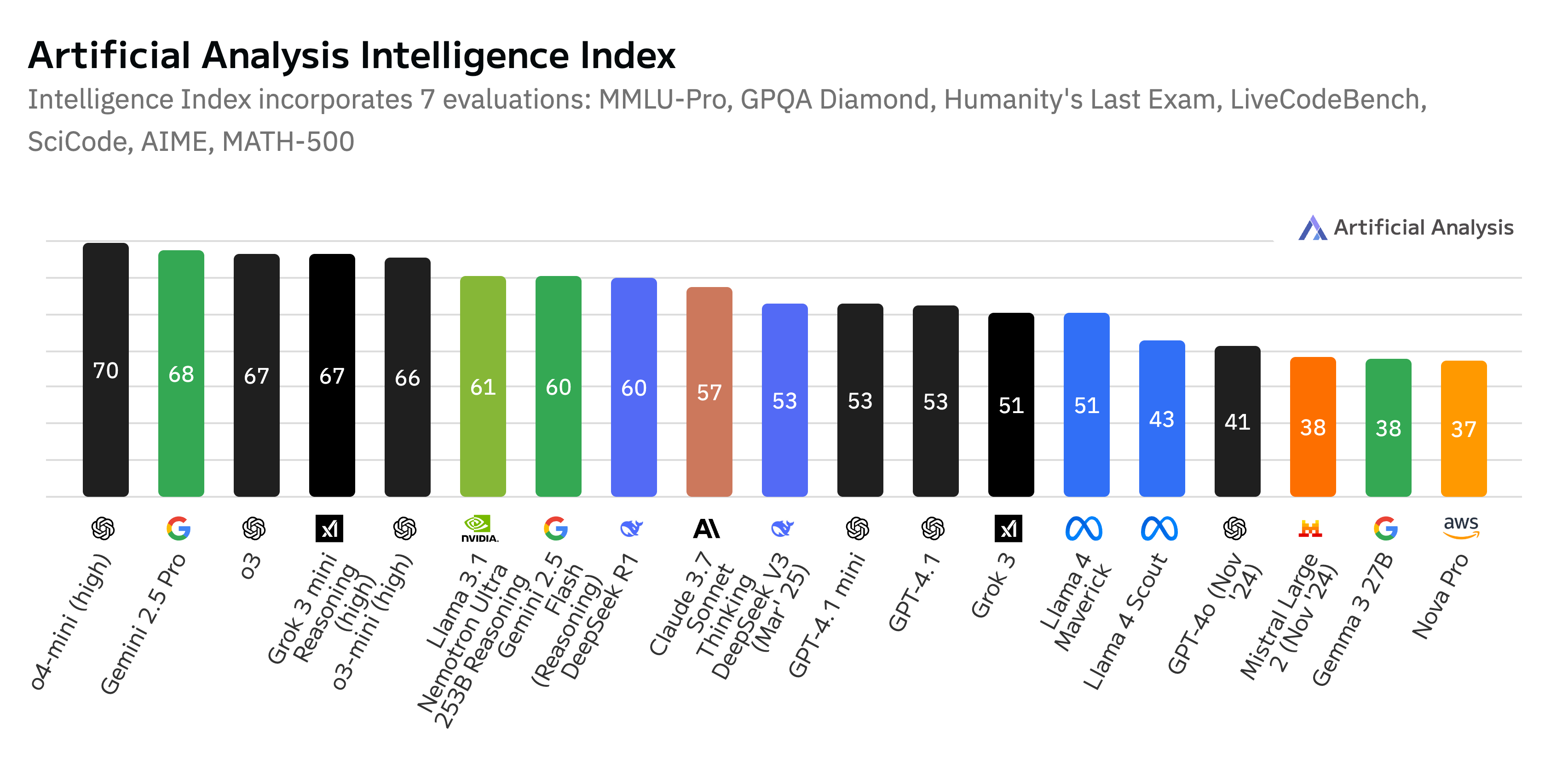}
   \caption{As of April 2025, our flagship model \ultra is the most ``intelligent'' open model according to \href{https://artificialanalysis.ai/}{Artificial Analysis}. }
   \label{fig:aa}
\end{figure}
\begin{figure}[h]
   \centering
   \includegraphics[width=\linewidth]{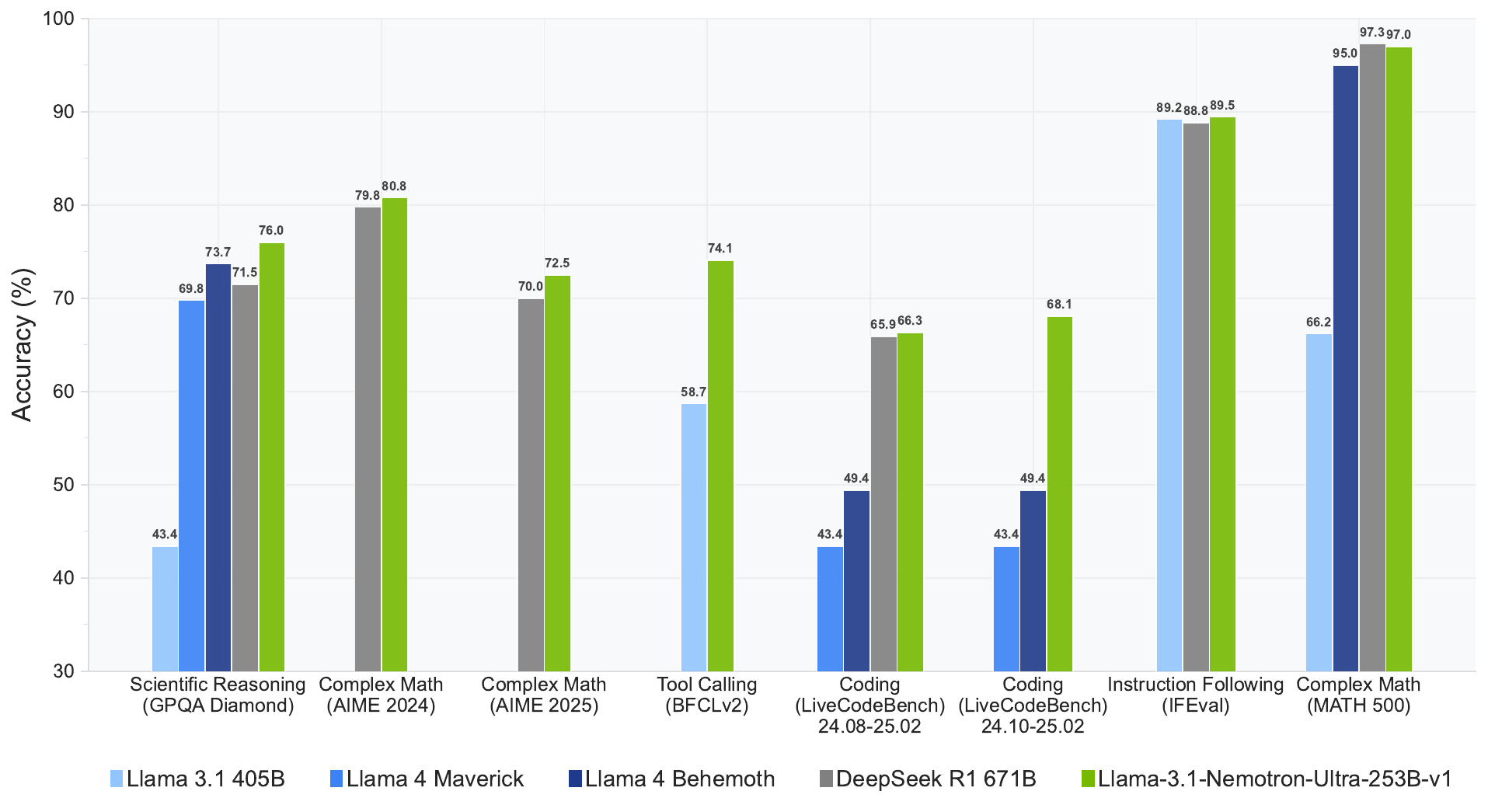}
   \caption{\ultra delivers leading performance among open models across a wide range of reasoning and non-reasoning benchmarks. }
   \label{fig:benchmark-bar}
\end{figure}

\section{Introduction}
In recent years the pace of language model development has been increasing, leading to rapid improvements in performance across a wide range of natural language processing tasks. Most recently, the introduction of reasoning models such as OpenAI o1 \citep{openai_o1} and DeepSeek-R1 \citep{deepseek_r1} has marked a new phase of advancement, resulting in models that can think deeply about problems before answering. A defining characteristic of these models is their long responses, often containing long chains of thought, self-verification, reflection, and backtracking. Such long responses enable them to achieve state-of-the-art performance across a wide variety of tasks, including PhD-level STEM questions and competition-level math problems.

As reasoning capabilities increasingly depend on scaling at inference time, it has become essential to design models that are efficient to run during inference. Inference efficiency is no longer just a deployment concern—it is now a core limiting factor for overall model intelligence and the viability of agentic pipelines. As such, maximizing inference efficiency is a primary optimization objective for these models. Beyond raw inference efficiency, it is equally critical to expose control over reasoning behavior to the end user. Not all queries benefit from detailed multi-step reasoning—such responses may be unnecessarily verbose or even counterproductive in certain contexts. Granting users the ability to toggle reasoning on or off ensures that inference resources are allocated judiciously and that response styles remain appropriate to the task \citep{anthropic2025claude37}.

In this paper, we detail the training of the Llama-Nemotron (LN) family of models, an open family of heterogeneous reasoning models that deliver exceptional reasoning capabilities, inference efficiency, and an open license for enterprise use. The models come in three sizes— \nano (8B), \super (49B), and \ultra (253B). Notably, \ultra outperforms DeepSeek-R1 while fitting on a single 8xH100 node and achieving higher inference throughput. These models are derived from Llama 3.1 and Llama 3.3 \citep{grattafiori2024llama}, and are optimized for high-throughput inference while delivering strong reasoning performance and a context length of 128K tokens. Each model supports a reasoning toggle that lets users dynamically switch between standard chat and reasoning modes at inference time using a lightweight system prompt: \texttt{"detailed thinking on/off"}. This design enables both cost-effective general-purpose use and detailed multi-step reasoning, without requiring separate models or architectures.

The Llama-Nemotron models are constructed in five stages. The first stage consists of optimizing inference efficiency with neural architecture search (NAS) from the Llama 3 series of models and applying Feed-Forward Network (FFN) Fusion. The second stage includes recovery training with knowledge distillation and continued pretraining. The third stage is supervised fine-tuning (SFT) on a mix of standard instruction data and reasoning traces from strong teachers such as DeepSeek-R1, which enables the model to perform multi-step reasoning. The fourth stage involves large-scale reinforcement learning on complex mathematics and STEM datasets, a crucial step for enabling the student model to surpass its teacher's capabilities. For \ultra{}, this phase yields a substantial performance boost on the GPQA-D benchmark, cementing it as the best open-source model for scientific reasoning. To enable such large-scale RL training, we develop a custom training framework that contains a number of optimizations, most notably generation in FP8. The final stage is a short alignment phase focused on instruction following and human preference.

As part of this release, we also open-source the \href{https://huggingface.co/datasets/nvidia/Llama-Nemotron-Post-Training-Dataset}{Llama-Nemotron-Post-Training-Dataset}, a carefully curated dataset used during the supervised and reinforcement learning stages of training for \nano, \super, and \ultra. It is designed to target key capabilities such as mathematical reasoning, coding, science, and instruction following, and consists of synthetic responses generated by a range of open-source models. Prompts and responses are filtered for quality, correctness, and complexity to provide strong training signals across a diverse set of tasks. 

According to Artificial Analysis (shown in Figure~\ref{fig:aa}), an independent benchmarking and analysis company focused on evaluating artificial intelligence models and API providers, \ultra is the most intelligent open-sourced model as of April 2025.
This release represents one of the largest contributions to the open source community in support of developing reasoning models.

\section{Creating Inference-Optimized Models}
\label{sec:nas}

The \super{} and \ultra{} models are optimized for efficient inference using the \emph{Puzzle} framework~\citep{bercovich2024puzzledistillationbasednasinferenceoptimized}. Puzzle is a neural architecture search (NAS) framework that transforms large language models into hardware-efficient variants under real-world deployment constraints, as illustrated in Figure~\ref{fig:puzzle_overview}. Starting from a Llama 3 Instruct model (Llama~3.3-70B-Instruct for \super{} and Llama~3.1-405B-Instruct for \ultra{}), Puzzle applies \textit{block-wise local distillation} to build a library of alternative transformer blocks. Each block is trained independently and in parallel to approximate the function of its parent block while improving computational properties such as latency, memory usage, or throughput. 
This process allows each alternative block to approximate the original behavior with a certain accuracy-efficiency tradeoff profile; that is, some blocks in the library are more efficient but may incur some quality degradation—introducing an explicit tradeoff between computational cost and model accuracy. The block variants include:
\begin{itemize}
    \item \textbf{Attention removal}: Some blocks omit the attention mechanism entirely, reducing both compute and KV-cache memory consumption.
    \item \textbf{Variable FFN dimensions}: The feed-forward network's intermediate size is varied, enabling compression at different granularity levels (e.g., 87\%, 75\%, 50\%, down to 10\% of the original hidden size).
\end{itemize}
While Puzzle supports additional operations—including grouped-query attention (GQA) \citep{gqa} with different numbers of key-value heads, linear alternatives to attention, and no-op substitutions—empirical evaluation showed that attention removal and FFN compression were the most effective for optimizing the \super{} and \ultra{} models in terms of overall throughput and memory savings.

\begin{figure}[t]
\label{fig:puzzle_overview}
     \centering
     \includegraphics[width=\linewidth]{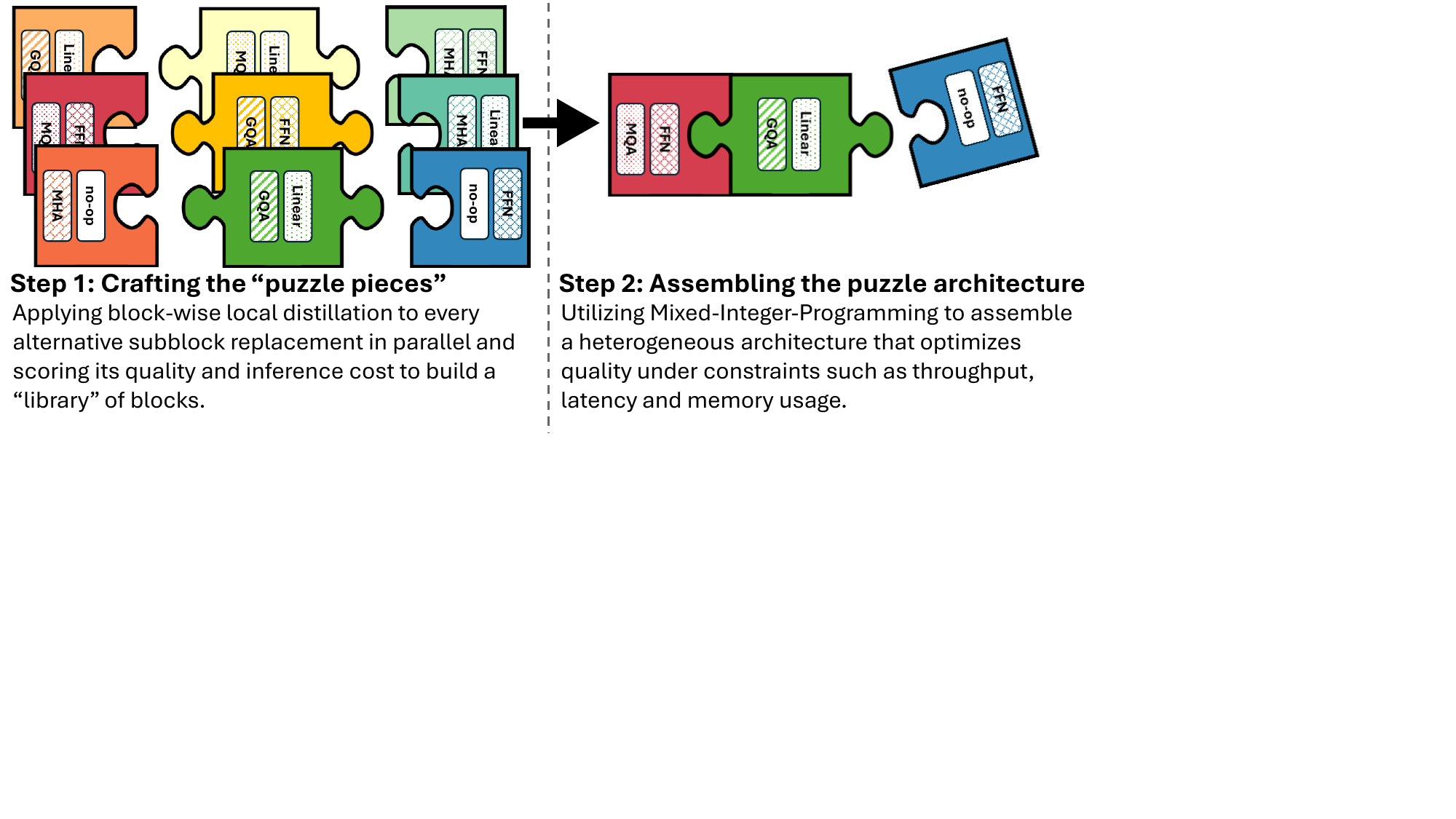}
     \caption{Overview of the Puzzle framework.}
     \label{fig:puzzle_overview}
 \end{figure}

Once the block library is built, Puzzle assembles a complete model by selecting one block per layer. This selection is governed by a mixed-integer programming (MIP) solver that identifies the most efficient configuration under a given set of constraints, such as hardware compatibility, maximum allowed latency, total memory budget, or desired inference throughput. Because Puzzle supports multiple block variants per layer with different accuracy-efficiency tradeoff profiles, it enables users to precisely target any point on the accuracy-efficiency Pareto frontier. For example, Puzzle can generate models that meet specific constraints relevant to agentic systems or deployment pipelines -- such as bounded memory use or tight end-to-end response time.

\paragraph{Vertical Compression with FFN Fusion.} For the \ultra{} model, we introduce an additional compression technique called \textit{FFN Fusion}~\citep{bercovich2025ffnfusionrethinkingsequential}, designed to reduce sequential depth and improve inference latency. This technique leverages a structural property that emerges after Puzzle removes some attention layers: the model often contains consecutive FFN blocks. FFN Fusion identifies such sequences and replaces them with fewer, wider FFN layers that can be executed in parallel. This reduces the number of sequential steps without compromising expressivity, and significantly improves compute utilization—especially on multi-GPU setups where inter-layer communication overhead is non-negligible.

\subsection{Deployment Constraints and Efficiency Targets}

\paragraph{\super} is optimized to run efficiently on a single NVIDIA H100 GPU with tensor parallelism 1 (TP1). Using Puzzle, we produce a model that achieves a $5\times$ throughput speedup over Llama~3.3-70B-Instruct at batch size 256 and TP1. With one H100 GPU, even when Llama~3.3-70B-Instruct is run at its optimal configuration with TP4, \super{} at TP1 still delivers a $\geq$2.17$\times$ throughput advantage. The model is also optimized under a constraint of approximately 300K cached tokens (batch~size~$\times$~sequence~length), measured at FP8 precision on a single H100 GPU. For instance, this corresponds to processing batch size 16 and sequence length 18,750.

\paragraph{\ultra} is optimized for a full H100 node (8 GPUs). During Puzzle’s architecture search phase, the model is constrained to achieve at least a 1.5$\times$ latency reduction over Llama~3.1-405B-Instruct. After applying FFN Fusion, the final model achieves a 1.71$\times$ latency improvement. \ultra{} is also optimized under cached tokens constraints, supporting up to 3M tokens at FP8 precision and 600K tokens at BF16 precision on an H100 node.

\begin{figure}[t]
\label{fig:acc_throughput}
     \centering
     \includegraphics[width=1\linewidth]{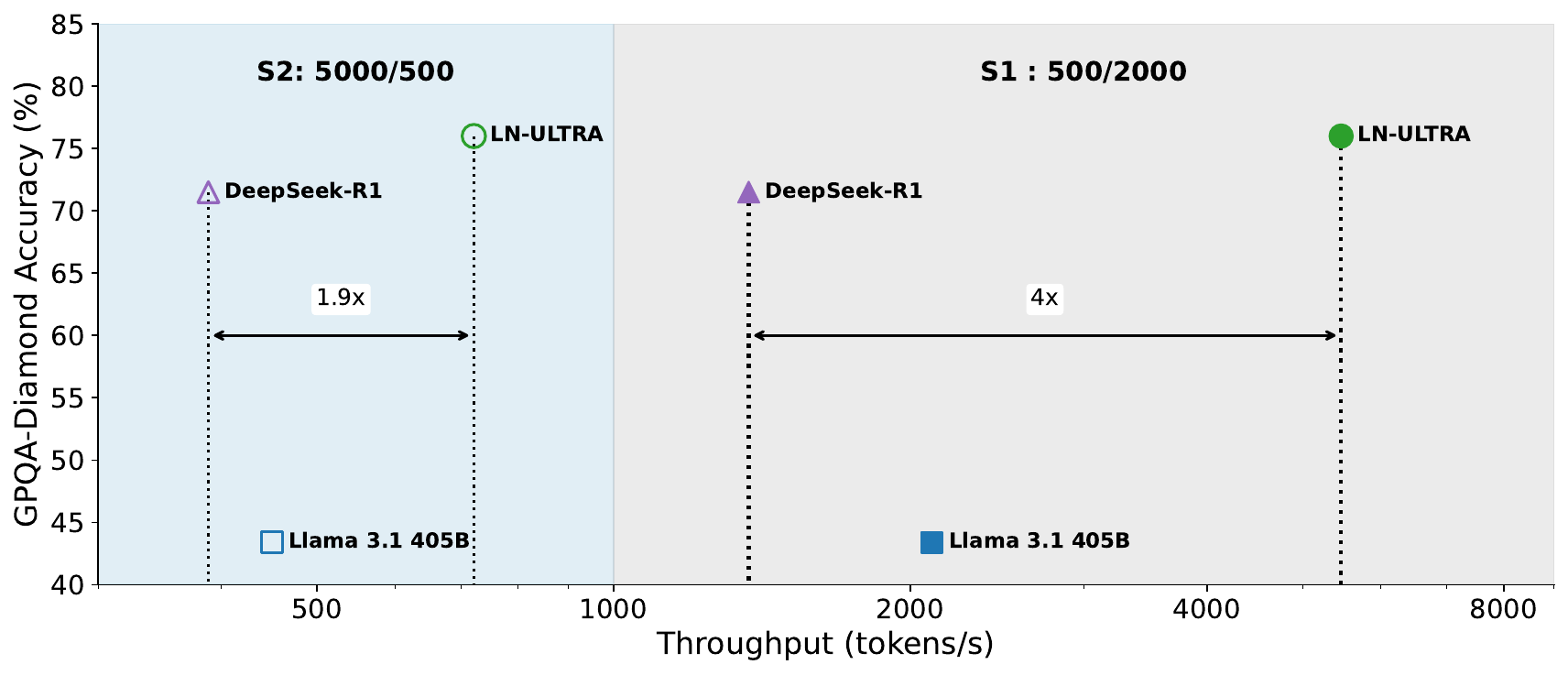}
     \caption{GPQA-Diamond Accuracy vs. Throughput. We measure on two settings, S1: 500/2000 (ISL/OSL); S2: 5000/500 (ISL/OSL). Both with 250 concurrent users. Models are served with FP8. Note that we use 8$\times$H100 for \ultra and Llama 3.1 405B, but 8$\times$H200 for Deepseek-R1 because of its size.}
     \label{fig:acc_throughput}
 \end{figure}

Figure~\ref{fig:acc_throughput} illustrates the trade-off between GPQA-Diamond accuracy (\%) and processing throughput (tokens/s) for two settings. Notably, \ultra consistently outperforms DeepSeek-R1 and Llama-3.1-405B in both accuracy and efficiency across these settings, clearly positioning it as a superior choice on the accuracy-throughput Pareto curve.

\subsection{Post-NAS Training: Knowledge Distillation and Continued Pretraining}

Following the NAS phase, both \super{} and \ultra{} undergo additional training to improve inter-block compatibility and recover any quality loss introduced during blockwise substitution.

\begin{itemize}
    \item \super{} is trained for 40B tokens using a knowledge distillation objective over the \textit{Distillation Mix} dataset introduced by~\cite{bercovich2024puzzledistillationbasednasinferenceoptimized}.
    \item \ultra{} is first trained with knowledge distillation for 65B tokens using the same distillation dataset, followed by 88B tokens of continued training on the Nemotron-H phase 4 pretraining dataset~\citep{nvidia2025nemotronhfamilyaccurateefficient}.
\end{itemize}

This final pretraining step allows \ultra{} to not only match but surpass the reference model Llama~3.1-405B-Instruct in key benchmarks, demonstrating that aggressive architecture optimization can be reconciled with high model performance through short distillation and pretraining (see Table~\ref{tab:ultra-cpt-table}).

\begin{table}[h]\small
\centering
\begin{tabular}{@{}lccc@{}}
\toprule
\textbf{Task} &
  \textbf{\begin{tabular}[c]{@{}c@{}}\ultra \\CPT{} \end{tabular}} &
  \textbf{\begin{tabular}[c]{@{}c@{}}Llama-3.3 \\70B-Instruct \end{tabular}} &
  \textbf{\begin{tabular}[c]{@{}c@{}}Llama-3.1 \\405B-Instruct\end{tabular}}  \\ \toprule
MMLU            & 88.1                 & 81.4  &  \textbf{88.6} \\
MATH500         & \textbf{80.4}        & 73.6  & 69.6  \\
HumanEval       &  \textbf{88.4}	   & 84.1  & 86.0  \\
RULER 128K      & \textbf{83.2}        &  52.2 & 73.7  \\
\bottomrule
\end{tabular}
\caption{Comparison of \ultra after the continued pretraining phase (before supervised and reinforcement learning) to Llama 3 models.}
\label{tab:ultra-cpt-table}
\end{table}

\section{Synthetic Data}
\label{sec:data}
We curate both reasoning and non-reasoning data for supervised fine-tuning. For reasoning samples, we include the system instruction \texttt{"detailed thinking on"}, and for non-reasoning samples, we use \texttt{"detailed thinking off"}. This setup allows the model to learn to toggle reasoning behavior at inference time based on the prompt. Below, we describe our focused data curation process for each mode.

\subsection{Reasoning on}
\label{sec:CT_reasoning_on}

\subsubsection{Math}
To construct the math reasoning portion of our data we used a pipeline described by~\cite{moshkov2025aimo2winningsolutionbuilding}. A high-level overview of this pipeline is provided below, with full details available in the original publication.

We collect a large set of mathematical problems from \href{https://artofproblemsolving.com/community}{Art of Problem Solving (AoPS) community forums}. We include all forum discussions except ``Middle School Math'' which was found to be too easy and unhelpful for training in our early experiments. After retrieving forum discussions we perform the following steps to extract problems and synthesize new solutions. We use Qwen2.5-32B-Instruct~\citep{qwen2.5} for all steps in the pipeline unless noted otherwise.

\vspace{-0.25in}
\paragraph{Problem Extraction:} We prompt an LLM to identify and extract all problems from the initial forum posts. While most posts contain a single problem, some include multiple problems or none at all.

\vspace{-0.25in}
\paragraph{Problem Classification:} Each extracted problem is classified into the following categories:
\begin{itemize}
    \item Proof problem or not
    \item Multiple choice question or not
    \item Binary question (yes-or-no answer) or not
    \item Valid problem or not. For example, problems that are lacking context or referring to other problems are considered invalid.
\end{itemize}
We remove all proof problems, multiple-choice questions, binary questions, and invalid problems from the final dataset.

\vspace{-0.25in}
\paragraph{Answer Extraction:} We extract the final answer from forum discussions, without attempting to extract full solutions. Only the final answer expression is extracted to enable automatic correctness checking.

\vspace{-0.25in}
\paragraph{Benchmark Decontamination:} Following~\cite{yang2023rethinking} we use an LLM-based comparison to remove questions that closely resemble those in popular math benchmarks.

\vspace{-0.25in}
\paragraph{Solution generation:} We prompt DeepSeek-R1~\citep{deepseek_r1} and Qwen2.5-Math-7B-Instruct~\citep{qwen2.5} to solve each problem multiple times producing ``reasoning'' and ``non-reasoning'' solutions respectively. We use 16 generations per problem for DeepSeek-R1 and 64 generations per problem for Qwen2.5-Math-7B-Instruct.

\vspace{-0.25in}
\paragraph{Solution Filtering:} As the final filtering step, we remove any solutions that do not reach the expected answer. Predicted and expected answers are compared by prompting Qwen2.5-32B-Instruct~\citep{qwen2.5} to judge their equivalence in the context of the problem. For problems where the final answer cannot be extracted, we treat the most common answer across all available solution candidates as the ground truth.

All prompts and scripts necessary to run the above pipeline are available in \href{https://nvidia.github.io/NeMo-Skills/openmathreasoning1/dataset/}{NeMo-Skills}. 

\subsubsection{Code}

The code reasoning dataset is constructed via a multi-stage process involving question collection, solution generation, and post-processing steps, as described by \cite{ahmad2025opencodereasoning}.

\vspace{-0.25in}
\paragraph{Question Collection and Verification:}
We aggregate 28,904 unique competitive programming questions from diverse sources including TACO \citep{li2023taco}, APPS \citep{apps2021}, CodeContests \citep{li2022codecontests}, and CodeForces \citep{openr1_codeforces}, after performing exact-match deduplication. To ensure evaluation integrity against benchmarks like \citep{jain2025livecodebench, li2022codecontests, chen2021codex_humaneval, austin2021mbpp}, we rigorously check for contamination using the method from \cite{yang2023rethinking}. This involves cosine similarity checks and semantic evaluation by LLM judges (Llama-3.3-70B \citep{grattafiori2024llama}, Qwen2.5-32B \citep{qwen2.5}). Manual verification confirms negligible overlap ($<0.3\%$), validating the question set.

\vspace{-0.25in}
\paragraph{Solution Generation:}
We employ DeepSeek-R1 \citep{deepseek_r1} to generate multiple solutions per question, primarily in Python, with C++ solutions also generated for specific benchmark testing \citep{openr1_ioi}. Solutions are generated using Nucleus Sampling \citep{holtzman2020nucleus_sampling} (temperature 0.6, top-p 0.95) via SGLang \citep{zheng2024sglang}, explicitly prompting for reasoning steps enclosed in \textit{\textless think\textgreater} tags.

\vspace{-0.25in}
\paragraph{Post-Processing and Refinement:}
We refine generated responses by verifying the presence of reasoning traces, extracting solution code segments (demarcated by \verb|python...|), removing samples with code inside reasoning tags, and validating syntax using Tree Sitter~\citep{tree_sitter}. This process yields approximately 488K Python samples.

\vspace{-0.25in}
\paragraph{Data Scaling Insights:}
While some studies suggest small datasets suffice for inducing reasoning \citep{openr1, s1_simpletesttimescaling, bespoke_stratos, openthoughts}, especially in mathematics, our experiments indicate large-scale data is crucial for high performance on coding benchmarks. An ablation study scaling the dataset from 25k to 736k samples showed continuous improvement. Initial scaling (25k-100k) provides gains, but focusing generation on harder problems from CodeContests before expanding to the full question set yields the most significant performance boosts. The scaling curve does not plateau, emphasizing the importance of large, diverse, and challenging problem sets for advancing code generation capabilities, suggesting a need for methods to create or source more difficult problems at scale.

\subsubsection{Science}

We curate a diverse set of open-ended and multiple-choice questions (MCQs) from both in-house and external sources. These include question-answer pairs extracted from StackOverflow~\citep{stack_exchange} and synthetically generated MCQ questions.

\vspace{-0.25in}
\paragraph{Synthetic Question Generation:}
To create synthetic questions, we define a broad set of academic topics (e.g., physics, biology, chemistry) and their subtopics using Nemotron-4-340B-Instruct~\citep{nvidia2024nemotron4340btechnicalreport}. We specify multiple difficulty levels to ensure a diverse and scalable dataset. We prompt Qwen2.5 models~\citep{qwen2.5} to generate MCQs conditioned on the topic, subtopic, and difficulty level. Each question is verified for format compliance. Following the OpenMathInstruct-2~\citep{toshniwal2024openmathinstruct} pipeline, we augment the dataset by prompting Qwen2.5 to generate variations of the original questions.

\vspace{-0.25in}
\paragraph{Benchmark Decontamination:}
To ensure fair evaluation, we perform decontamination on the entire set of questions—both real and synthetic—against the test sets of major science benchmarks such as GPQA~\citep{rein2023gpqa}, MMLU~\citep{hendrycks2021measuringmassivemultitasklanguage}, and MMLU-Pro~\citep{wang2024mmluprorobustchallengingmultitask}, following the approach outlined in~\cite{yang2023rethinking}.

\vspace{-0.25in}
\paragraph{Solution Generation:}
For all questions in the dataset, we use DeepSeek-R1~\citep{deepseek_r1} to generate multiple reasoning traces. For questions without ground-truth answers, we apply majority voting across generated solutions to infer the most likely correct answer.

\subsubsection{General}
For general domain data, we follow the generation pipeline established in \citet{nvidia2024nemotron4340btechnicalreport}. We generate synthetic prompts covering various tasks such as open QA, closed QA, extraction, and brainstorming. We also source real-world user prompts from publicly available datasets with permissive licenses. For responses, we prompt DeepSeek-R1~\citep{deepseek_r1} for multiple generations and perform rejection sampling using the Llama-3.1-Nemotron-70B reward model~\citep{nemotron70breward}. This ensures that the responses are of high quality.

\subsection{Reasoning off}
\label{sec:CT_reasoning_off}

To train the model to follow the reasoning toggle instruction, we construct paired data where each prompt has both a reasoning response and a non-reasoning response. Specifically, we randomly sample prompts from the reasoning dataset in Section \ref{sec:CT_reasoning_on} and generate corresponding non-reasoning responses using Llama-3.1-Nemotron-70B-Instruct~\citep{nemotron70binstruct} for general domain prompts and Llama-3.3-70B-Instruct for others. Each response is tagged with the appropriate system instruction—\texttt{“detailed thinking on”} for reasoning and \texttt{“detailed thinking off”} for non-reasoning. This pairing enables the model to learn to modulate its reasoning behavior based on the system prompt.

Responses are then filtered according to ground truth answers or reward models. We also leverage public permissive datasets on function calling and safety, augmenting them to train the model and improve its capabilities in these areas. To further improve performance on general tasks, we use a feedback-edit system, described in Section~\ref{sec:its}.

\subsubsection{General-Domain Open-ended Inference-Time Scaling} \label{sec:its}

To generate high-quality general-domain open-ended responses, we employ Llama-3.1-Nemotron-70B-Instruct \citep{nemotron70binstruct} in conjunction with a novel Feedback-Edit Inference-Time-Scaling system, described by \citet{wang2025dedicatedfeedbackeditmodels}. The process begins with 20k first-turn prompts sourced from ShareGPT \citep{sharegpt2023} and WildChat-1M \citep{zhao2024wildchat1mchatgptinteraction}. We use Llama-3.1-Nemotron-70B-Instruct to generate multiple initial responses for each prompt. These responses are refined through a three-stage process: a dedicated Feedback model identifies areas for improvement, a dedicated Edit model makes targeted edits based on the feedback, and a dedicated Select model chooses the best edited response. The resulting dataset comprises 20k first-turn prompts and their corresponding high-quality responses.

\begin{table}[h]\small
\centering
\begin{tabular}{@{}lrr@{}}
\toprule
\textbf{Domain / Split}               & \textbf{Samples} & \textbf{\% of total} \\ 
\toprule
\textbf{Math}                           & 22,066,397 & 66.8\%  \\
\quad Reasoning \textit{on}    & 2,225,427  & 6.7\%  \\
\quad Reasoning \textit{off}   & 19,840,970 & 60.1\%  \\
\midrule
\textbf{Code}                           & 10,108,883 & 30.6\% \\
\quad Reasoning \textit{on}    & 991,706  & 3.0\%   \\
\quad Reasoning \textit{off}   & 9,117,177  & 27.6\%\\[2pt]
\midrule
\textbf{Science}                        & 708,920  & 2.1\%   \\
\quad Reasoning \textit{on}    & 708,920  & 2.1\%   \\
\quad Reasoning \textit{off}   & 0             & 0.0\%   \\[2pt]
\midrule
\textbf{Chat}                           & 39,792 & 0.12\% \\
\quad Reasoning \textit{on}    & 8,574  & 0.03\%   \\
\quad Reasoning \textit{off}   & 31,218             & 0.09\%   \\
\midrule
\textbf{Instruction Following  }        & 56,339   & 0.17\%  \\
\textbf{Safety  }                       & 31,426   & 0.10\%  \\ \midrule
\textbf{Total}                 & 33,011,757 & 100\%   \\ 
\bottomrule
\end{tabular}
\caption{Synthetic data by domain with reasoning splits. }
\end{table}

\section{Supervised Fine-Tuning}
\label{sec:sft}

Supervised fine-tuning (SFT) plays a critical role in transferring reasoning capabilities into the Llama-Nemotron models. 
While prior stages such as NAS and CPT focus on architectural efficiency and broad knowledge transfer, SFT helps distill reasoning behavior from strong teacher models like DeepSeek-R1~\citep{deepseek_r1} by training on task-specific reasoning traces. It also establishes fine-grained control over response style using the \texttt{"detailed thinking on/off"} instruction. Recent studies~\citep{deepseek_r1, openthoughts, bespoke_stratos, lightr1} have shown that this reasoning SFT can substantially improve performance on complex reasoning tasks. Our results confirm these findings, highlighting the importance of training on large-scale, high-quality reasoning traces during SFT for eliciting robust reasoning abilities in downstream usage. This section builds upon the synthetic data described in Section~\ref{sec:data} and provides further implementation details specific to each model.

\subsection{General Methodology}
All models are trained using a token-level cross-entropy loss over the instruction-tuning data. For most settings, training batches mix reasoning and non-reasoning data, where prompts are paired with responses conditioned on the respective system instruction— \texttt{"detailed thinking on/off"}.

We observe that models require higher learning rates to effectively learn from long reasoning traces, especially due to sequence-length-dependent token loss averaging. Extended training over multiple epochs improves performance, particularly for smaller models, a trend also observed in prior work~\citep{lightr1}. We use Adam optimizer for training all models. Using a cosine learning rate decay with linear warmup to around 10\% of total steps helps with stability of training, which was crucial for \ultra{}.

\subsection{Model-Specific Training}

\paragraph{\nano{}} differently from other models below, undergoes a three-stage SFT pipeline using a global batch size of 256 using sequence packing with effective sequence length of 32k tokens. In the first stage, the model is fine-tuned exclusively on reasoning data from code, math, and science domains (Section~\ref{sec:CT_reasoning_on}) with a learning rate of $1\mathrm{e}{-4}$ for four epochs. This prevents failure modes such as repetitive completions. In the second stage, we introduce non-reasoning data (Section~\ref{sec:CT_reasoning_off}) mixed with reasoning samples, allowing the model to learn reasoning control. In the final stage, a smaller blend focused on chat, instruction-following, and tool-calling is used.

\paragraph{\super{}} is trained on the full SFT dataset for a single epoch using a fixed learning rate of $5\mathrm{e}{-6}$, sequence length of 16k and a global batch size of 256. Smaller-scale runs suggested that performance improves up to 3–4 epochs with larger learning rates ($5\mathrm{e}{-5}$), but training was constrained by computational and time limits. Recent works~\citep{lightr1} show that rejection fine-tuning can further improve performance; however, it does not yield gains in our experiments and is therefore omitted.

\paragraph{\ultra{}} is trained on the full dataset using sequence packing with effective sequence length of 24k and a global batch size of 256 to maximize token throughput—an essential strategy when fine-tuning large models with long-context reasoning data. Initial ablation runs indicated that higher learning rates such as $5\mathrm{e}{-5}$ generally improve outcomes, but consistently high learning rates caused training instability, including gradient explosions. To mitigate this, we implement a linear warmup to $1\mathrm{e}{-5}$, followed by cosine decay to $1\mathrm{e}{-6}$ with a warmup ratio of 10\%. Despite these measures, training encountered gradient explosions and numerical instability after the first epoch. This required training resumption with reinitialized optimizer states, after which successful convergence was achieved.

\section{RL for Reasoning}
\label{sec:reasoning-rl}
As described in Section~\ref{sec:sft} and illustrated in Table~\ref{tab:ultra-table}, models can acquire strong capabilities through supervised fine-tuning by distilling knowledge from powerful teachers. However, distillation inherently sets an upper bound on the student's performance, particularly when the student's base model is more capable than the teacher's. Using supervised fine-tuning, \ultra{} can approach the performance of DeepSeek-R1 but not exceed it. To enable student to surpass their teacher, large-scale reinforcement learning is an essential approach, as it allows the model to continually explore new possibilities and engage in self-learning. Consistent with the findings of \citet{deepseek_r1}, our preliminary experiments indicate that reinforcement learning (RL) yields suboptimal results for smaller models compared to distillation. Given these observations and resource constraints, we apply reasoning RL exclusively to \ultra{}, enabling it to surpass its teacher model and setting a new state-of-the-art on GPQA amoung open models. 

\subsection{Training Procedure}
\begin{figure}[h]
    \centering
    \includegraphics[width=0.8\linewidth]{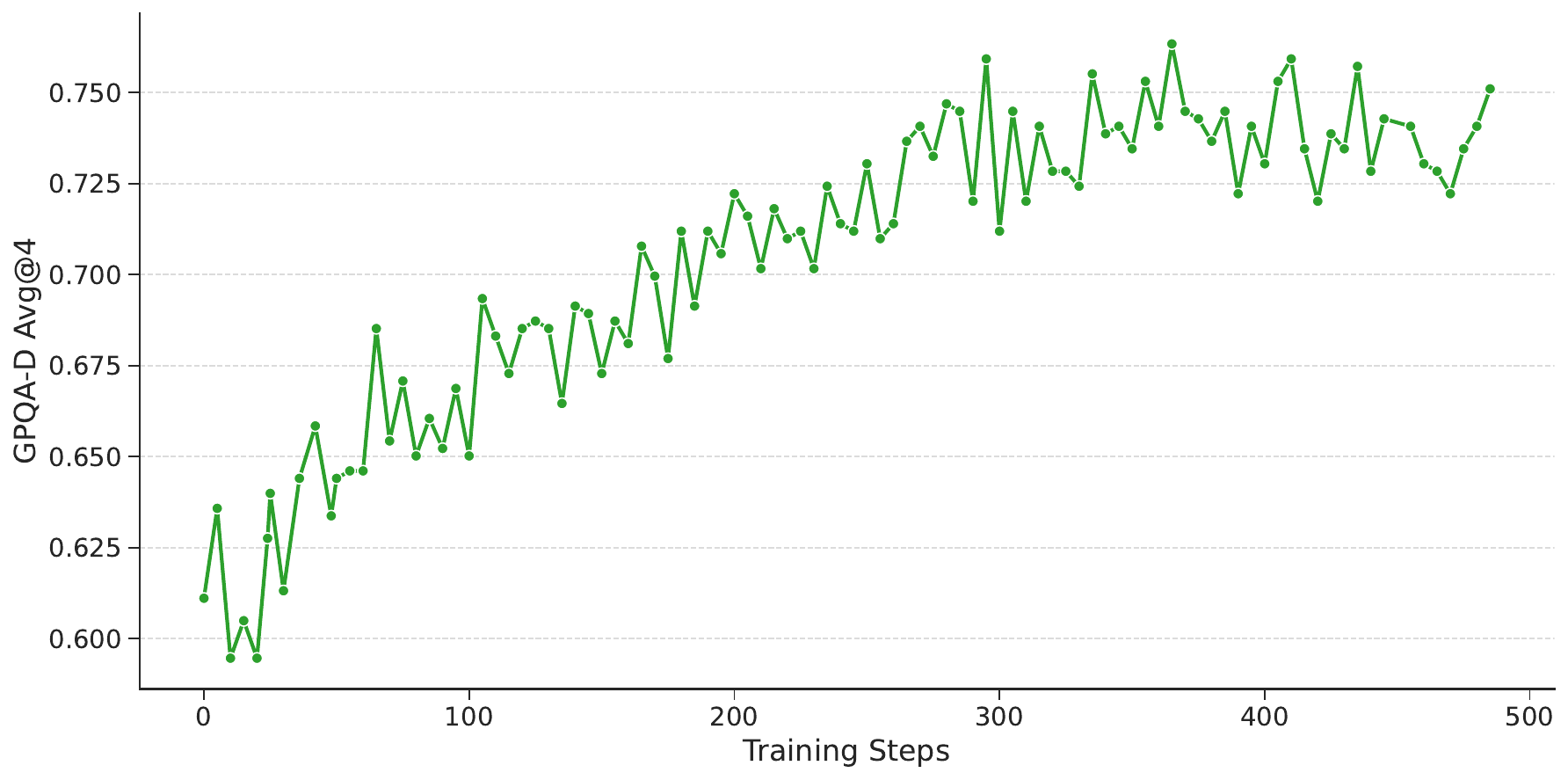}
    \caption{Accuracy on GPQA-Diamond throughout the reasoning RL training for \ultra}
    \label{fig:whole_reasoning_rl}
\end{figure}

For \ultra{}, we enhance the model's scientific reasoning capabilities through large-scale reinforcement learning, leveraging the Group Relative Policy Optimization (GRPO) algorithm~\citep{deepseek-math}. We use a rollout prompt size of 72 and sample 16 responses per prompt with $temperature=1$ and $top\_p=1$. During training, we set global batch size as 576 and conduct 2 gradient updates per rollout. We train our model until it achieves convergence on reasoning tasks. Figure~\ref{fig:whole_reasoning_rl} shows the accuracy score on GPQA-Diamond as our training progresses. With our optimized training infrastructure (see Section~\ref{sec:infra}), the whole training takes approximately 140k H100 hours.

In this training phase, we leverage two types of rewards:

\textbf{Accuracy rewards:} For each training example, a ground truth answer (a number, a sentence, or a paragraph) is provided. We serve the Llama-3.3-70B-Instruct model to judge whether the policy's predictions match the ground truth answer.

\textbf{Format rewards:} Following \citet{deepseek_r1}, we employ a format reward to ensure the model puts its thinking process between "<think>" and "</think>" tags when using \texttt{"detailed thinking on"} mode. We also check for the non-existence of thinking tags when using \texttt{"detailed thinking off"} mode.

\begin{figure}[h]
    \centering
    \includegraphics[width=0.8\linewidth]{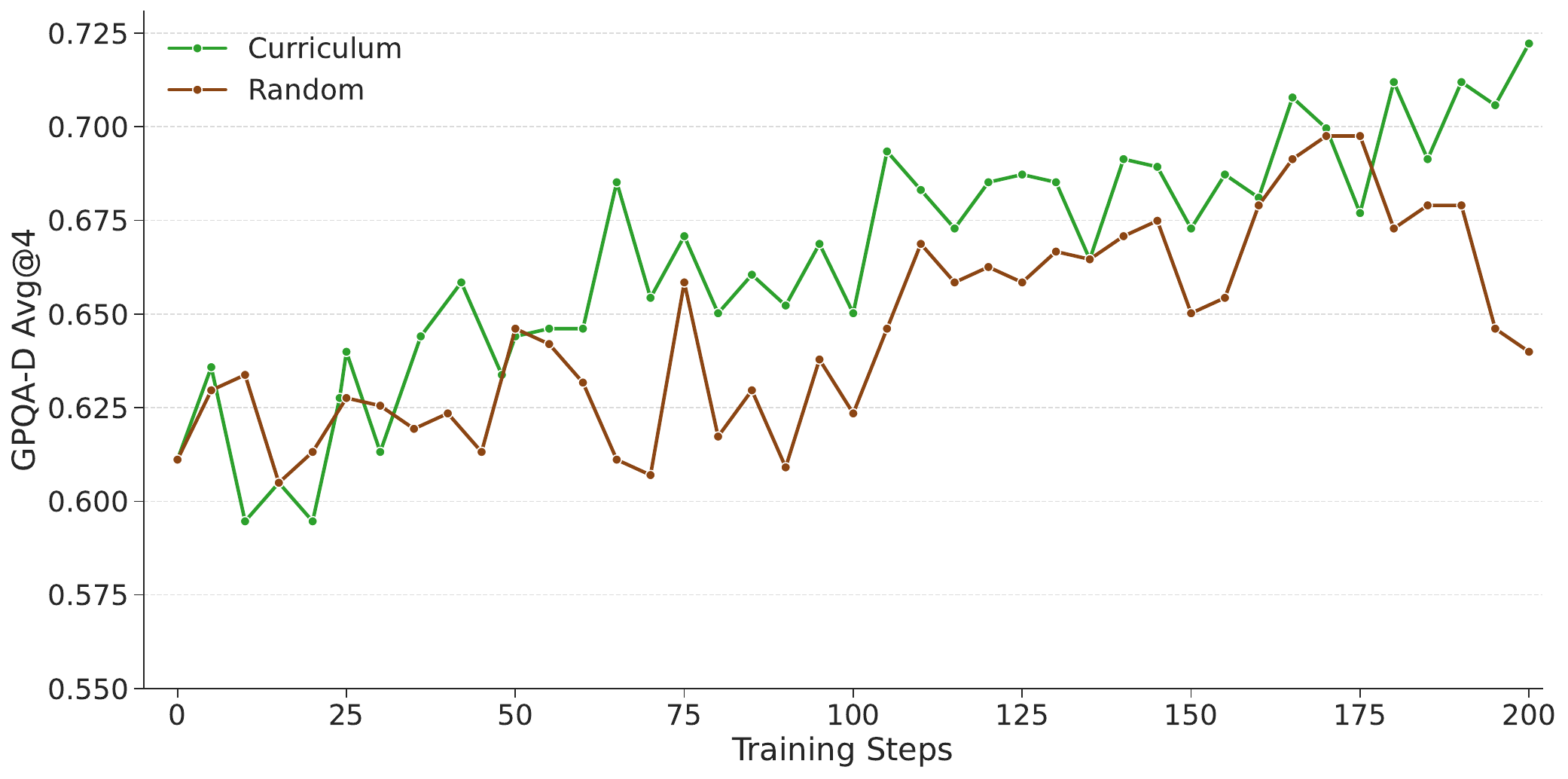}
    \caption{Ablation on curriculum vs non-curriculum.}
    \label{fig:curriculum_ablation}
\end{figure}

To ensure that the model is adequately challenged, we preprocess the data by independently generating 8 responses per question using \super{}, calculating the pass rate, and then intentionally discarding prompts with a pass rate of 0.75 or higher, thereby increasing the difficulty of the training data. Besides data filtering, we also find curriculum training to be helpful, as it allows the model to gradually learn from a progression of tasks with increasing difficulty. 
Specifically, we implement a progressive batching strategy leveraging pre-calculated pass rate as a difficulty metric. Given a fixed batch size, the core of our approach involves dynamically calculating a target distribution of pass rates for each sequential batch. This distribution is modeled using a Gaussian function centered on a difficulty level that progresses from high pass rates (easier examples) for initial batches to low pass rates (harder examples) for later batches. Samples are allocated to each batch primarily based on this target distribution, considering the available count for each pass rate, with any remaining batch capacity filled by prioritizing pass rates with the largest remaining sample pools. This ensures a controlled, gradual increase in average sample difficulty across batches, while samples inside a batch are randomly shuffled. Figure~\ref{fig:curriculum_ablation} demonstrates the effectiveness of our curriculum strategy, which stabilizes the training process and achieves higher accuracy.

\subsection{Infrastructure}
\label{sec:infra}
\subsubsection{Overview}
\label{sec:infra_overview}
We primarily use NeMo-Aligner~\citep{nemo-aligner} to perform RL training, where we use a development branch that implements GRPO and heterogeneous model support. We implements the generation
stage using vLLM~\citep{vllm} and the training stage using Megatron-LM~\citep{megatronlm}. The training and inference stages are co-located on the same GPUs. 

The total GPU count used was 72 nodes of 8xH100. The training model parallelism used was: tensor parallel=8 with sequence parallel, context parallel=2, pipeline parallel=18, and data parallel=2. The generation model parallelism was tensor parallel=8,
and data parallel=72. The details of how this parallelization strategy is chosen is explained in \ref{sec:infra_memory}. Generation was performed in FP8, and training in BF16 with FP32 optimizer states. 

Each stage maintains its own set of model
weights, which are synced at the start of each step. First, all training weights are all-gathered over the training pipeline parallel dimension, converted into vLLM format, and written into shared memory. Then all training stage memory is released or offloaded to host. Next, vLLM is awoken from
sleep mode, loads the newly saved model weights from shared memory, and begins generating. After
generations have finished, vLLM GPU memory is released using sleep mode=2, and all training memory is reloaded. 

\subsubsection{Memory Profiling and Optimizations}
\label{sec:infra_memory}

One of the major challenges in enabling the GRPO training of the \ultra{} is memory management.
The training jobs are scheduled to a shared cluster environment. In the cluster, each node has 
8 H100 GPUs, dual socket 32-core CPUs, and 2TB CPU DRAM. On the other hand, the model in BF16 
type takes $253 \times 2 \approx 500GB$ memory. Moreover, as mentioned in Section \ref{sec:infra_overview}, 
in order to improve GPU utilization,
we determine to stack training and inference stages on the same set of nodes. 
Without careful memory management, it is very easy to
encounter out-of-memory errors in both GPU and CPU memory allocations.

In order to better track the memory usage over the course of training, we have developed three
simple memory profiling tools to monitor the memory usage: GPU memory utilization using
PyTorch, CPU memory utilization using psutil, and \emph{/dev/shm} utilization using the df command. The GPU/CPU memory profilers
help us track the GPU/CPU memory usage at different code pointers. The \emph{/dev/shm} profiler is needed
as we use \emph{/dev/shm} to pass the weights from the trainer to the vLLM server, and the host is 
configured to allocate up to 1TB for the \emph{/dev/shm} space.

With the help of these profiling tools, we are able to pinpoint the specific memory allocations that cause
out-of-memory errors, and then design solutions to overcome the issues. The first challenge is
weight preparation. When we all-gather training weights across pipeline parallel stages, we have encountered
extremely big tensors due to the heterogeneous architecture. One of the tensors has 13B elements, and occupies 26B
GPU memory in the BF16 type. We need to release unused GPU memory periodically, and move some of the tensor conversion operations
to CPU in order to control the GPU memory usage in this stage. The second challenge is the vLLM GPU memory utilization.
With tensor parallel equal to 8, we expect each GPU to keep $500 / 8 \approx 62GB$ from the weights in BF16. Considering KV cache,
activations, and GPU memory occupied by the trainer, we have a very tight budget for the vLLM. We have to disable the cudagraph
feature to avoid GPU out-of-memory in vLLM. However, when we enable FP8 inference generation as explained in \ref{sec:fp8},  GPU memory
budget becomes a lot looser, and we get to enable the cudagraph feature again. The final challenge is the GPU and CPU memory
usage in the trainer. Tensor parallelism = 8 is a natural choice to partition the full model into the 8xH100 GPUs available in the same node. As the model architecture is heterogeneous, we need to insert identity layers in order to balance
the pipeline stages in the pipeline parallelism. We want to have enough pipeline parallelism to avoid training OOM in GPU 
and checkpoint saving OOM in CPU. On the other hand, we also want to reduce the number of pipeline stages to reduce the
communication costs. With all of the trade-offs, we find that the best pipeline parallelism setting is 18.
The activations also consume a lot of memory, and we need to keep 18 micro-batches in the case when pipeline parallelism is 18.
We end up using context parallel = 2 and sequence parallel to reduce the activation memory consumption 
to prevent GPU OOM in training. With all these tuning, we finally choose tensor parallel=8 with sequence parallel, context parallel=2, pipeline parallel=18, and data parallel=2 to achieve $> 90\%$ utilization of the GPUs while avoiding
any hosts from encountering GPU or CPU out-of-memory errors.

\subsubsection{FP8 Inference Generation}
\label{sec:fp8}
We identify the generation stage as the dominant component of the step time. In order to improve performance, we implement a path to support the use of vLLM's online FP8 generation mode, which executes all GEMMs in FP8 using per token activation scaling factors and per tensor weight scaling factors. We implement custom vLLM weight loaders capable of loading BF16 weights supplied by the training stage, and casting to FP8 weights and scaling factors at runtime. Because vLLM does not support directly initializing models in FP8, we also implement meta-weight tensor initialization to avoid materializing the full BF16 inference engine, which would cause an out-of-memory error in the GPU.

In all, we observe a peak FP8 generation throughput of 32 tokens/s/GPU/prompt, a 1.8x generation speedup against BF16, and to our knowledge the highest decoding throughput observed in reasoning training at this scale. We observe a 1.4x speedup from FP8 generation alone, and an additional 0.4x from the reduction in memory usage, which allows us to enable vLLM's cudagraph feature.

\section{RL for Preference Optimization}
\label{sec:alignment}

\subsection{Instruction Following}
\label{sec:alignment_IF}
After training for scientific reasoning, we do a short RL run optimizing instruction following capabilities for the \super{} and \ultra{}. We use a similar verification setup as \citet{zhou2023instruction}, and generate synthetic instruction following prompts that contain from one to ten detailed instructions. We run RL  with the RLOO algorithm \citep{ahmadian2024back} for less than 120 steps using our instruction following verifier as a reward, with a batch size of 128 prompts. We find such training boosts performance on conventional instruction following benchmarks as well as reasoning benchmarks. 
\vspace{-0.08in}
\subsection{RLHF}
We use RLHF to improve the model on general helpfulness and chat capabilities while carefully maintaining its proficiency in other areas. As shown in Table~\ref{tab:super-table}, \super{}, a 49B model, achieves an Arena Hard score of 88.3, beating proprietary models such as Claude 3.5 Sonnet and GPT-4o-2024-05-13 as well as much larger open models such as Llama-3.1-405b-instruct and Mistral-large-2407.

To achieve this, we use iterative online RPO~\citep{nvidia2024nemotron4340btechnicalreport,sun2025rewardawarepreferenceoptimizationunified} to maximize the reward predicted by Llama-3.1-Nemotron-70B-Reward~\citep{nemotron70breward} over prompts from HelpSteer2~\citep{helpsteer2}. For each iteration, we use a learning rate $\alpha$ of 4e-7, KL penalty $\beta$ of 1e-5, reward scale $\eta$ of 3.0, and batch size of 64, training for 500 steps. Two iterations of online RPO increase the Arena Hard score from 69.1 to 88.1. More interestingly, this process also improves the model's performance on all other adopted benchmarks except IFEval. Since neither the dataset nor the reward model is optimized for math, coding, science, or function calling, we speculate that RLHF helps the model better utilize its existing knowledge and skills.

We follow the same process for \ultra{}, except that GRPO is employed. For each prompt, we sample 8 responses. We train the model for 30 steps, using a learning rate of 3e-7, batch size of 288, and KL penalty $\beta$ of 1e-3.

For \nano, we conduct two rounds of offline RPO with on-policy data. We use a mixture of reasoning and non-reasoning data with appropriate system prompts in the first round of RPO to improve reasoning control, followed by a second round with on-policy generations targeting instruction following improvements. For each RPO round we train up to 400 steps with a learning rate $\alpha$ of 7e-7, KL penalty $\beta$ of 3e-2, and batch size of 512.

\section{Evaluations on Reasoning and Chat Benchmarks}

\subsection{Benchmarks}
We evaluate all Llama-Nemotron models across two benchmark categories: reasoning and non-reasoning.

\vspace{2pt}
\noindent\textbf{Reasoning Benchmarks.} These include the American Invitational Mathematics Examination for years 2024 (AIME24) and 2025(AIME25), GPQA-Diamond~\citep{rein2024gpqa}, LiveCodeBench~\citep{jain2024livecodebench}, and MATH500~\citep{lightman2023lets}. AIME25 is split into two parts: AIME25-I and AIME25-II, each containing 15 problems. For \nano, we use AIME25-I only; for \super and \ultra, we evaluate on the full 30-question set. As AIME25 was released recently, it is less likely to overlap with training data. Thus, stronger performance on this benchmark is indicative of better generalization, especially on math problems outside the training distribution. LiveCodeBench contains questions indexed by date, and we report results on two specific ranges—(2408–2502) and (2410–2502)—to enable fair comparison with previously reported baselines. 

\vspace{2pt}
\noindent\textbf{Non-Reasoning Benchmarks.} These include IFEval(Strict-Instruction)~\citep{zhou2023instruction} for instruction following, BFCL V2 Live~\citep{berkeley-function-calling-leaderboard} for tool use via function calling, and Arena-Hard~\citep{arenahard2024} for evaluating alignment with human conversational preferences.

All evaluations are conducted at 32k context length, even though training was performed with a maximum context length of 16k for \super and 24k for \ultra. We observed consistent improvements when evaluating at expanded context lengths, as shorter limits can truncate long reasoning traces and lead to incomplete generations—particularly on benchmarks that require multi-step reasoning. We use temperature 0.6 and top-p 0.95 for reasoning-on evaluations, and temperature 0 (greedy decoding) for reasoning-off.  We generate up to 16 completions per prompt and report average pass@1 accuracy. Checkpoints are selected based on performance on a subset of reasoning benchmarks. As observed in prior works ~\citep{moshkov2025aimo2winningsolutionbuilding}, evaluation on reasoning-heavy tasks such as AIME can exhibit high variance due to small dataset size and generation randomness. Reported numbers may vary across repeated runs or sampling strategies.

\subsection{\nano}
Table~\ref{tab:nano-table} shows that \nano achieves strong performance across all reasoning benchmarks, including AIME25-I and LiveCodeBench, despite its small size. This demonstrates the effectiveness of our SFT pipeline and curated reasoning datasets in transferring structured reasoning to compact models. For \nano, carefully balancing data-distribution across math, coding, and stem areas has been important to achieve near state-of-the-art accuracies at the SFT stage. For example, our early experiments showed worse accuracies especially in chemistry related questions, one of the major areas in GPQA-D. Upsampling chemistry related data samples in the STEM subset of the overall SFT blend helped to achieve higher GPQA-D accuracies. The RPO stages at the end of the post-training pipeline mainly targeted IFEval accuracy improvement as shown in Table~\ref{tab:nano-table}.

\newcolumntype{V}{D{|}{\;\vert\;}{-1}}

\begin{table}[t]\small
\centering
\begin{tabular}{@{}l V V c c c@{}}
\toprule
\textbf{Task} &
  \multicolumn{1}{c}{\scriptsize\textbf{\begin{tabular}[c]{@{}c@{}}\textsf{LN-Nano-SFT} \\ Reasoning\end{tabular}}} &
  \multicolumn{1}{c}{\scriptsize\textbf{\begin{tabular}[c]{@{}c@{}}\nano{} \\ Reasoning\end{tabular}}} &
  \scriptsize\textbf{\begin{tabular}[c]{@{}c@{}}DeepSeek-R1\\Distilled-Llama-8B\end{tabular}} &
  \scriptsize\textbf{\begin{tabular}[c]{@{}c@{}}Llama-3.1\\8B-Instruct\end{tabular}} &
  \scriptsize\textbf{\begin{tabular}[c]{@{}c@{}}DeepSeek-R1\\Distilled\end{tabular}} \\
  & \scriptsize\textbf{\textit{on}} | \scriptsize\textbf{\textit{off}} & \scriptsize\textbf{\textit{on}} | \scriptsize\textbf{\textit{off}} & & & \scriptsize\textbf{Qwen-7B}\\
\midrule
GPQA-Diamond  & 53.5|{33.3} & \textbf{54.1}|{39.4}  & 49.0 & 25.3  & 49.1 \\
AIME24        & \textbf{62.5}|{3.3} & 61.3|{3.0}  & 50.4 & 10.0  & 55.6 \\
AIME25-I      & \textbf{51.6}|{6.6} & 47.1|{0.0}  & 40.0 & 10.0  & 41.7 \\
MATH500       & 94.4|{38.0} & \textbf{95.4}|{36.6}  & 89.1 & 50.4  & 92.8 \\
BFCL V2 Live  & 62.9|{62.6} & \textbf{63.9}|{63.6}  & 37.8 & 44.3  & 39.2 \\
LiveCodeBench \scriptsize(2408–2502) & {-}|{-} & \textbf{46.6}|{-}  & 39.6 & 11.8  & 37.6 \\
IFEval & 69.9|{69.9} & {79.29}|\textbf{82.1}  & 73.4 & 81.8  & 67.6 \\
\bottomrule
\end{tabular}
\caption{\nano{} and \textsf{LN-Nano-SFT} versus comparably sized models, split by Reasoning mode.}
\label{tab:nano-table}
\end{table}

\subsection{\super}
Table~\ref{tab:super-table} compares \super to other models in its weight class where it performs competitively across both reasoning and non-reasoning tasks. In reasoning-off mode, \super performs on par with Llama-3.3-70B, the model it based on. In reasoning-on mode, it outperforms competing models such as DeepSeek-R1-Distilled-Llama-70B, providing strong reasoning capabilities without sacrificing instruction following. These results show that this single model offers the strengths of both reasoning-optimized and non-reasoning models, making it effective for general assistant and structured reasoning use cases.
Additionally, as shown in Table~\ref{tab:super-table}, reasoning-focused SFT causes a noticeable drop in IFEval scores. To recover instruction-following capabilities, we apply a dedicated IFEval RL run (see Section~\ref{sec:alignment_IF}) to ensure that strong reasoning does not come at the cost of degraded general assistant behavior.
Our experimental results reveal another trade-off: optimizing for instruction following (as measured by IFEval) can compromise conversationality (as measured by Arena-Hard), and conversely, prioritizing conversationality may detract from instruction following performance. To address this, we applied model merging to \super, selecting a checkpoint from the Pareto frontier that balances these objectives. Due to mixed outcomes, we did not adopt this approach for other models. The only area where \super underperforms is on LiveCodeBench, which is attributable to its SFT phase being conducted on an earlier version of the dataset, unlike \nano and \ultra. We plan to address this and improve coding-related reasoning performance in a future model refresh.

\newcolumntype{V}{D{|}{\;\vert\;}{-1}}

\begin{table}[t]\small
\centering
\begin{tabular}{@{}l V V c c c@{}}
\toprule
\textbf{Task} &
    \multicolumn{1}{c}{\scriptsize \textbf{\begin{tabular}[c]{@{}c@{}}\textsf{LN-Super-SFT} \\ Reasoning\end{tabular}}} &
    \multicolumn{1}{c}{\scriptsize \textbf{\begin{tabular}[c]{@{}c@{}}\super{} \\ Reasoning\end{tabular}}} &
    \scriptsize \textbf{\begin{tabular}[c]{@{}c@{}}DeepSeek-R1-\\Distilled-Llama-70B\end{tabular}} &
    \scriptsize\textbf{\begin{tabular}[c]{@{}c@{}}QwQ-32B\end{tabular}} &
    \scriptsize\textbf{\begin{tabular}[c]{@{}c@{}}Llama-3.3 \\ 70B-Instruct\end{tabular}} \\
    & \scriptsize\textbf{\textit{on}} | \scriptsize\textbf{\textit{off}} & \scriptsize\textbf{\textit{on}} | \scriptsize\textbf{\textit{off}} & & & \\
\midrule
GPQA-Diamond  & 63.8|{46.6} & \textbf{66.7}|{50.0}  & 65.2 & 58.8 & 50.5 \\
AIME24        & 63.3|{17.5} & 67.5|{16.7}           & 70.0 & \textbf{79.5} & 25.8 \\
AIME25        & 50.0|{6.7} & 60.0|{16.7}           & 55.0 & \textbf{65.8} & 6.7 \\
MATH500       & 93.2|{76.8} & \textbf{96.6}|{74.0}  & 94.5 & 96.2 & 73.8 \\
BFCL V2 Live  & 73.3|{62.5} & {73.7}|\textbf{73.9}  & 65.5    & 71.6 & 60.4 \\
LiveCodeBench \scriptsize{(2408–2502)} & 40.9|{28.7} & 45.5|{29.7}  & 57.5 & \textbf{63.4} & - \\
IFEval        & 81.9|{83.0} & \textbf{89.2}|{89.0}  & 85.1 & 86.3 & 92.1 \\
Arena Hard    & {-}|{-}  & 88.3|{-}           & 65.4 & \textbf{90.5} & 72.9 \\
\bottomrule
\end{tabular}
\caption{\super{} versus comparably sized models, split by Reasoning mode.}
\label{tab:super-table}
\end{table}

\subsection{\ultra}
Table~\ref{tab:ultra-table} and Figure~\ref{fig:benchmark-bar} show that \ultra matches or outperforms all existing open-weight models across reasoning and non-reasoning benchmarks. It achieves state-of-the-art performance on GPQA among open models, demonstrating the efficacy of our large-scale reinforcement learning training. Unlike prior state-of-the-art models such as DeepSeek-R1, which require 8$\times$H200, \ultra is optimized to run efficiently on a single 8$\times$H100 node, offering improved inference throughput and deployment efficiency.

From Table~\ref{tab:ultra-table}, we observe that the \textsf{LN-Ultra-SFT} model approaches the performance of DeepSeek-R1 on several reasoning benchmarks, including GPQA and AIME. However, the RL stage is critical for surpassing DeepSeek-R1, particularly on GPQA. This highlights the complementary strengths of SFT and RL: SFT builds a strong foundation by distilling reasoning behavior from teacher models, while RL is essential for surpassing teacher performance and further enhancing reasoning capabilities.

We also find that there is a trade-off between the extent of SFT training and the success likelihood of subsequent RL. Although we had access to SFT checkpoints with higher benchmark scores, we initialized RL from an earlier checkpoint to improve final RL outcomes.

\newcolumntype{V}{D{|}{\;\vert\;}{-1}}

\begin{table}[h]\small
\centering
\begin{tabular}{@{}l V V c c c c@{}}
\toprule
\textbf{Task}
  & \multicolumn{1}{c}{\scriptsize \textbf{\begin{tabular}[c]{@{}c@{}}\textsf{LN-Ultra-SFT} \\ Reasoning\end{tabular}}}
  & \multicolumn{1}{c}{\scriptsize \textbf{\begin{tabular}[c]{@{}c@{}}\ultra{} \\ Reasoning\end{tabular}}}
  & \scriptsize \textbf{\begin{tabular}[c]{@{}c@{}}DeepSeek \\ R1\end{tabular}}
  & \scriptsize \textbf{\begin{tabular}[c]{@{}c@{}}Llama-4 \\ Behemoth\end{tabular}}
  & \scriptsize \textbf{\begin{tabular}[c]{@{}c@{}}Llama-4 \\ Maverick\end{tabular}}
  & \scriptsize \textbf{\begin{tabular}[c]{@{}c@{}}Llama-3.1 \\ 405B-Instruct\end{tabular}} \\
  & \scriptsize\textbf{\textit{on}} | \scriptsize\textbf{\textit{off}} & \scriptsize\textbf{\textit{on}} | \scriptsize\textbf{\textit{off}} & & & & \\
\midrule
GPQA-Diamond              & 66.4|{46.0} & \textbf{76.0}|{56.6}  & 71.5 & 73.7 & 69.8 & 43.4 \\
AIME24                   & 74.6|{46.7} & \textbf{80.8}|{20.0}  & 79.8 & –    & –    & 20.0    \\
AIME25                   & 60.4|{16.7} & \textbf{72.5}|{16.7}  & 70.0 & –    & –    & 0.0    \\
MATH500                  & 96.6|{84.4} & 97.0|{80.4}  & \textbf{97.3} & 95.0 & –    & 66.2    \\
BFCL V2 Live             & {74.6}|\textbf{74.9} & 74.1|{73.6}  & –    & –    & –    & 58.7 \\
LiveCodeBench \scriptsize{(2408–2502)}& 60.6|{30.1} & \textbf{66.3}|{29.0}  & 65.9 & –    & –    & –    \\
LiveCodeBench \scriptsize{(2410–2502)}& 61.8|{-} & \textbf{68.1}|{-}  & –    & 49.4 & 43.4 & –    \\
IFEval                   & 83.2|{79.4} & {88.9
}|\textbf{89.5}  & 88.8 & –    & –    & 89.2 \\
Arena Hard               & {-}|{-}     & 87.0|{-}  & \textbf{92.0} & –    & –    & 66.2 \\
\bottomrule
\end{tabular}
\caption{\ultra{} versus the strongest open-weight models, split by reasoning mode.}
\label{tab:ultra-table}
\end{table}

\section{Evaluations on Judging Capability}
In addition to reasoning and chat capabilities where the models are trained for, we evaluate our models on an out-of-distribution task, LLM-as-a-Judge, to further assess their performance. Specifically, we test them on JudgeBench~\citep{tan2024judgebench}, where the task is to differentiate between high-quality and low-quality responses. As shown in Table~\ref{tab:Judgebench}, our models outperform top proprietary and open-source models. Notably, \ultra{} emerges as the best open-source model, significantly surpassing DeepSeek-R1 and trailing only behind o3-mini(high). Furthermore, \super{} also outperforms o1-mini, demonstrating that our models exhibit strong generalization capabilities across diverse tasks.

\begin{table}[h]\small
\centering
\begin{tabular}{@{}lccccc@{}}
\toprule
\textbf{Model} &
  \textbf{\begin{tabular}[c]{@{}c@{}}Knowledge\end{tabular}} &
  \textbf{\begin{tabular}[c]{@{}c@{}}Reasoning\end{tabular}} &
  \textbf{\begin{tabular}[c]{@{}c@{}}Math \end{tabular}} &
  \textbf{\begin{tabular}[c]{@{}c@{}}Coding \end{tabular}} &
  \textbf{\begin{tabular}[c]{@{}c@{}}Overall \end{tabular}}  \\
  \toprule
o1-preview  & 66.23  & 79.59 & 85.71  & 85.71 & 75.43 \\
o1-mini  & 58.44  & 62.24 & 82.14 & 78.57 & 65.71  \\
o3-mini(low)       &  62.99	 &  69.39 & 83.93 & 83.33 & 70.57  \\
o3-mini(medium)          & 62.34  & 86.73  & 85.71 & 92.86 & 76.57  \\
o3-mini(high)          & 67.53 & \textbf{89.80} & 87.50 & \textbf{100.0} & \textbf{80.86}  \\
DeepSeek-R1         & 59.09  & 82.65 & 80.36 & 92.86 & 73.14  \\
\midrule
\super{} & 64.94 & 67.35 & 76.79 & 83.33 & 69.71 \\
\ultra{} & \textbf{70.13} & 81.63 & \textbf{89.29} & 92.86 & 79.14 \\
\bottomrule
\end{tabular}
\caption{Llama-Nemotron models demonstrate strong performance on JudgeBench.}
\label{tab:Judgebench}
\end{table}
\vspace{-2mm}

\section{Conclusions}
\label{sec:conclusions}
We present the Llama-Nemotron series of models. The models are released under a permissive license and we open-source the weights, training data, and code. The Llama-Nemotron series of models perform competitively with state-of-the-art reasoning models, while having low memory requirements and efficient inference capabilities.

We find that in the presence of a strong reasoning teacher, supervised fine-tuning on high-quality synthetic data generated by such teacher is very effective in adding reasoning capabilities to smaller models. 
However, to push reasoning capabilities beyond what is possible from a teacher reasoning model alone, it is necessary to run large-scale, curriculum-driven reinforcement learning from verifiable rewards training. 

We also show that to produce a great all-around model, e.g. a model which performs well on a wide variety of benchmarks, it is necessary to have several stages in the post-training pipeline.

\section*{Contributors}

We thank the following people for their invaluable contributions to the Llama-Nemotron effort.

\textbf{NAS, Distillation and Continued Pretraining}

Akhiad Bercovich*, Itay Levy*, Izik Golan*, Mohammad Dabbah*, Ran El-Yaniv*, Omri Puny, Ido Galil, Zach Moshe, Tomer Ronen, Najeeb Nabwani, Ido Shahaf, Oren Tropp, Ehud Karpas, Ran Zilberstein

\textbf{Post-training } 

Jiaqi Zeng*, Soumye Singhal*, Alexander Bukharin*, Yian Zhang*, Tugrul Konuk*, Gerald Shen*, Ameya Sunil Mahabaleshwarkar*, Bilal Kartal*, Yoshi Suhara*, Olivier Delalleau, Zijia Chen, Zhilin Wang, David Mosallanezhad, Adi Renduchintala, Haifeng Qian, Dima Rekesh, Fei Jia

\textbf{Data}

Somshubra Majumdar, Vahid Noroozi, Wasi Uddin Ahmad, Sean Narenthiran, Aleksander Ficek, Mehrzad Samadi, Jocelyn Huang, Siddhartha Jain, 
Igor Gitman, Ivan Moshkov, Wei Du, Shubham Toshniwal, George Armstrong, Branislav Kisacanin,
Matvei Novikov, Daria Gitman, Evelina Bakhturina, Jiaqi Zeng, Zhilin Wang, Tugrul Konuk, Ameya Sunil Mahabaleshwarkar, Bilal Kartal, Yoshi Suhara, Prasoon Varshney, Makesh Narsimhan, Jane Polak Scowcroft, John Kamalu, Dan Su, Kezhi Kong, Markus Kliegl, Rabeeh Karimi Mahabadi, Ying Lin, Sanjeev Satheesh, Jupinder Parmar, Pritam Gundecha, Brandon Norick, Joseph Jennings, Shrimai Prabhumoye, Syeda Nahida Akter, Mostofa Patwary, Abhinav Khattar, Deepak Narayanan, Roger Waleffe

\textbf{Infrastructure} 

Jimmy Zhang*, Bor-Yiing Su*, Guyue Huang*, Terry Kong, Parth Chadha, Sahil Jain, Christine Harvey, Elad Segal, Jining Huang, Sergey Kashirsky, Robert McQueen 

\textbf{Inference}

Izzy Putterman*, George Lam, Arun Venkatesan, Sherry Wu, Vinh Nguyen, Manoj Kilaru, Andrew Wang, Anna Warno, Abhilash Somasamudramath, Sandip Bhaskar, Maka Dong, Nave Assaf, Shahar Mor, Omer Ullman Argov, Scot Junkin, Oleksandr Romanenko, Pedro Larroy, Monika Katariya, Marco Rovinelli, Viji Balas, Nicholas Edelman, Anahita Bhiwandiwalla, Muthu Subramaniam, Smita Ithape, Karthik Ramamoorthy, Yuting Wu, Suguna Varshini Velury, Omri Almog, Joyjit Daw
 
\textbf{Evaluations and Safety}

Soumye Singhal*, Denys Fridman, Erick Galinkin, Michael Evans, Shaona Ghosh, Katherine Luna, Leon Derczynski, Nikki Pope, Eileen Long, Seth Schneider, Guillermo Siman, Tomasz Grzegorzek, Pablo Ribalta, Monika Katariya, Chris Alexiuk

\textbf{Program management} 

Joey Conway*, Ehud Karpas*, Trisha Saar*, Ann Guan, Krzysztof Pawelec, Shyamala Prayaga

\textbf{Leadership}

Tugrul Konuk*, Oleksii Kuchaiev*, Boris Ginsburg*, Oluwatobi Olabiyi*, Kari Briski*, Jonathan Cohen*, Bryan Catanzaro*, Jonah Alben, Yonatan Geifman, Eric Chung

* Core contributor

\newpage

\bibliography{references}
\bibliographystyle{references}

\end{document}